\documentclass[letterpaper, 10 pt, conference]{ieeeconf}
\IEEEoverridecommandlockouts 

\usepackage{microtype}

\usepackage{hyperref}
\hypersetup{colorlinks,linkcolor={green!50!black},citecolor={green!50!black},urlcolor={blue!80!black}}
\makeatletter \let\NAT@parse\undefined \makeatother
\usepackage[sort,compress]{cite}
\usepackage{graphicx} 
\usepackage{amsfonts}
\usepackage{amsmath,soul}
\usepackage{color}
\usepackage[font=small]{subcaption}
\usepackage{balance}
\usepackage[font=small]{caption}
\usepackage[linesnumbered,ruled,vlined]{algorithm2e}
\usepackage{multirow}

\DeclareMathOperator*{\argmin}{\arg\!\min}
\DeclareMathOperator*{\argmax}{\arg\!\max}
\usepackage{tabulary}
\newcolumntype{K}[1]{>{\centering\arraybackslash}p{#1}}

\title{\LARGE \bf Socially Aware Motion Planning with Deep Reinforcement Learning}

\author{Yu Fan Chen, Michael Everett, Miao Liu$^\dag$, and Jonathan P.\
  How
  \thanks{Laboratory of Information and Decision Systems,
    Massachusetts Institute of Technology, 77 Massachusetts Ave.,
    Cambridge, MA, USA\newline {\tt\footnotesize \{chenyuf2, mfe, jhow\}@mit.edu}}%
      \thanks{$^\dag$IBM Thomas J. Watson Research Center, Yorktown Heights, NY 10598, USA {\tt\footnotesize miao.liu1@ibm.com }}
}

\usepackage[svgnames]{xcolor} \definecolor{DarkGreen}{rgb}{0,0.5,0}
\definecolor{DarkRed}{rgb}{0.75,0,0}

\usepackage[authormarkuptext=name,addedmarkup=bf,authormarkupposition=left]{changes}
\definechangesauthor[name={S.~L.}, color={red}]{sl}
\definechangesauthor[name={J.~H.}, color={blue}]{jh}
\definechangesauthor[name={M.~L.}, color={purple}]{ml}
\definechangesauthor[name={J.~M.}, color={orange}]{jm}
\definechangesauthor[name={S.~C.}, color={DarkGreen}]{sc}
\definechangesauthor[name={M.~E.}, color={red}]{me}
\setremarkmarkup{(#2)}

\usepackage{tikz,mathtools}
\usepackage[capitalize]{cleveref}
\crefformat{equation}{(#2#1#3)}
\Crefformat{equation}{Equation~(#2#1#3)}
\Crefname{equation}{Equation}{Equations}

\usetikzlibrary{shapes,positioning,automata,arrows,fit,backgrounds,calc}
\tikzstyle{block} = [draw, fill=blue!20, rectangle,minimum height=1em,
minimum width=2em] \tikzstyle{sum} = [draw, fill=blue!20, circle, node
distance=1cm] \tikzstyle{input} = [coordinate] \tikzstyle{output} =
[coordinate] \tikzstyle{pinstyle} = [pin edge={to-,thin,black}]
\usetikzlibrary{trees} \usetikzlibrary{decorations.pathmorphing}
\usetikzlibrary{decorations.markings}
\definecolor{darkgreen}{rgb}{0,0.5,0}
\definecolor{darkred}{rgb}{220,20,60}

\makeatletter
\renewcommand\paragraph{\@startsection{subsubsection}{4}{\z@}%
{0.25ex \@plus.5ex \@minus.2ex}%
{-.15em}%
{\normalfont\normalsize\itshape}}
\makeatother

\begin{document}

\maketitle
\thispagestyle{empty} \pagestyle{empty}

\begin{abstract} 
For robotic vehicles to navigate safely and efficiently in pedestrian-rich environments, it is important to model subtle human behaviors and navigation rules (e.g., passing on the right). However, while instinctive to humans, socially compliant navigation is still difficult to quantify due to the stochasticity in people's behaviors. Existing works are mostly focused on using feature-matching techniques to describe and imitate human paths, but often do not generalize well since the feature values can vary from person to person, and even run to run. This work notes that while it is challenging to directly specify the details of what \textit{to} do (precise mechanisms of human navigation), it is straightforward to specify what \textit{not to} do (violations of social norms). Specifically, using deep reinforcement learning, this work develops a time-efficient navigation policy that respects common social norms. The proposed method is shown to enable fully autonomous navigation of a robotic vehicle moving at human walking speed in an environment with many pedestrians. 
\end{abstract}

\section{Introduction} \label{sec:intro}

Recent advances in sensing and computing technologies have spurred greater interest in various applications of autonomous ground vehicles. In particular, researchers have explored using robots to provide personal mobility services and luggage carrying support in complex, pedestrian-rich environments (e.g., airports and shopping malls)~\cite{bai_intention-aware_2015}. These tasks often require the robots to be capable of navigating efficiently and safely in close proximity of people, which is challenging because pedestrians tend to follow subtle social norms that are difficult to quantify, and pedestrians' intents (i.e., goals) are usually not known~\cite{kretzschmar_socially_2016}.

A common approach treats pedestrians as dynamic obstacles with simple kinematics, and employs specific reactive rules for avoiding collision~\cite{fox_dynamic_1997,van_den_berg_reciprocal_2008,phillips_sipp:_2011,berg_reciprocal_2011}. Since these methods do not capture human behaviors, they sometimes generate unsafe/unnatural movements, particularly when the robot operates near human walking speed~\cite{kretzschmar_socially_2016}. To address this issue, more sophisticated motion models have been proposed, which would reason about the nearby pedestrians' hidden intents to generate a set of predicted paths~\cite{kim_brvo:_2015,unhelkar_human-robot_2015}. Subsequently, classical path planning algorithms would be employed to generate a collision-free path for the robot. Yet, separating the navigation problem into disjoint prediction and planning steps can lead to the \emph{freezing robot problem}, in which the robot fails to find any feasible action because the predicted paths could mark a large portion of the space untraversable~\cite{trautman_robot_2015}.
A key to resolving this problem is to account for cooperation, that is, to model/anticipate the impact of the robot's motion on the nearby pedestrians. 

\begin{figure}[t]
	\centering
	\includegraphics [trim=0 0 0 0, clip, angle=0, width=0.8\columnwidth,
	keepaspectratio]{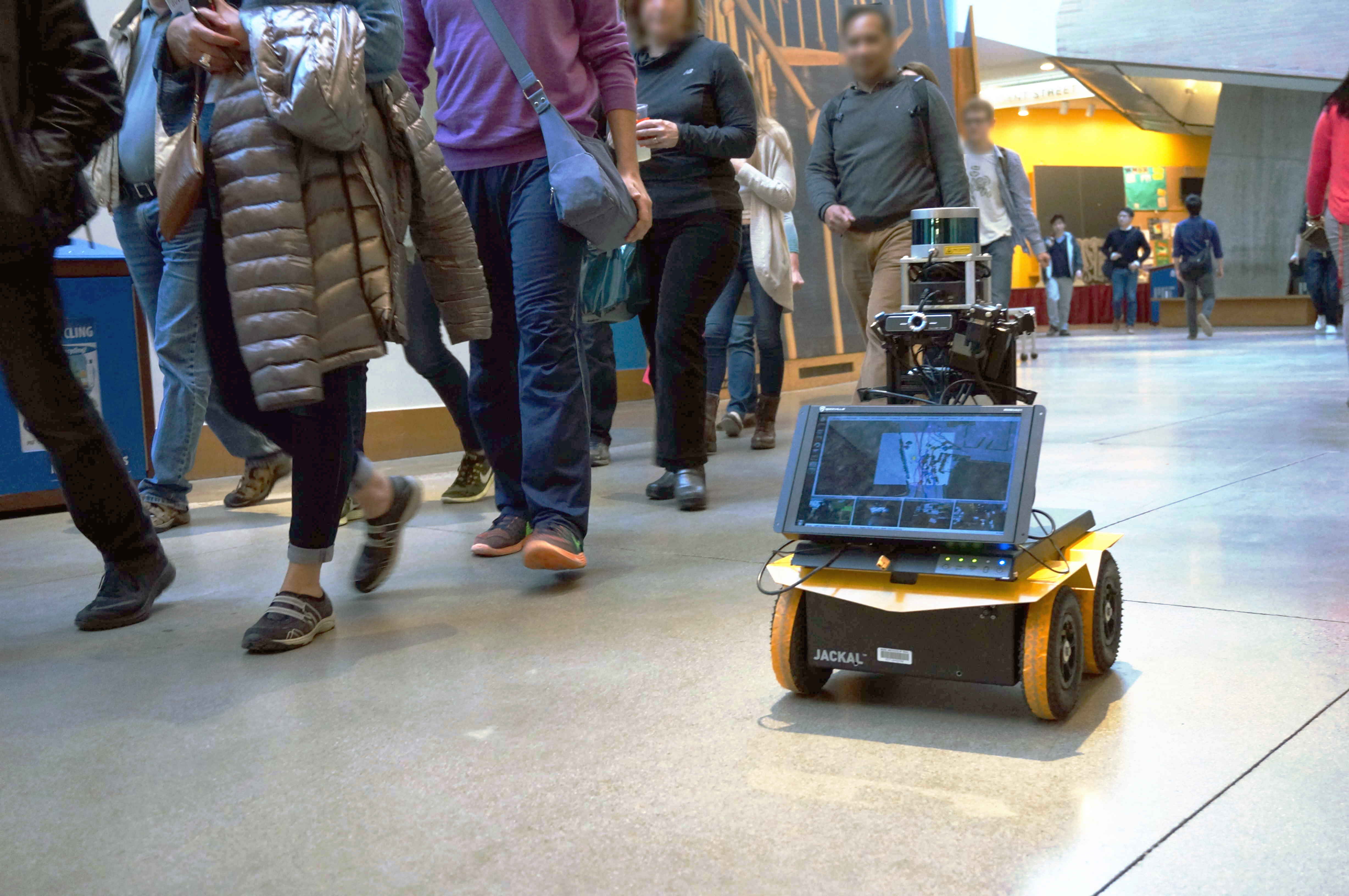}
	\caption{A robotic vehicle navigating autonomously in a pedestrian-rich environment. Accounting for social interactions is important for operating such vehicles safely and smoothly. 
	} 
	\label{fig:rover} 
\end{figure}

Existing work on cooperative, socially compliant navigation can be broadly classified into two categories, namely \textit{model-based} and \textit{learning-based}. Model-based approaches are typically extensions of multiagent collision avoidance algorithms, with additional parameters introduced to account for social interactions~\cite{helbing_social_1995,ferrer_robot_2013,ferrer_behavior_2014,kim_brvo:_2015,mehta_autonomous_2016}. For instance, to distinguish between human--human and human--robot interactions, the extended social forces model~\cite{ferrer_behavior_2014,ferrer_robot_2013} augments the potential field algorithm with additional terms that specify the repulsive forces (e.g., strength and range) governing each type of interaction. Model-based methods are designed to be computationally efficient as they often correspond to intuitive geometric relations; yet, it is unclear whether humans do follow such precise geometric rules. In particular, the force parameters often need to be tuned individually, and can vary significantly for different pedestrians~\cite{ferrer_behavior_2014}. Also, it has been observed that model-based methods can lead to oscillatory paths~\cite{chen_decentralized_2017,kretzschmar_socially_2016}. 

In comparison, learning-based approaches aim to develop a policy that emulates human behaviors by matching feature statistics, such as the minimum separation distance to pedestrians. In particular, Inverse Reinforcement Learning (IRL)~\cite{abbeel_apprenticeship_2004} has been applied to learn a cost function from human demonstration (teleoperation)~\cite{kim_socially_2015}, and a probability distribution over the set of joint trajectories with nearby pedestrians~\cite{kuderer_feature-based_2012,kretzschmar_socially_2016}. Compared with model-based approaches, learning-based methods have been shown to produce paths that more closely resemble human behaviors, but often at a much higher computational cost. This is because computing/matching trajectory features often requires anticipating the joint paths of all nearby pedestrians~\cite{kretzschmar_socially_2016}, and might depend on some  unobservable information (e.g., pedestrians' goals). 
More importantly, since human behaviors are inherently stochastic, 
the feature statistics calculated on pedestrians' paths can vary significantly from person to person, and even run to run for the same scenario~\cite{kim_socially_2015,kretzschmar_socially_2016}. This raises concerns over whether such feature-matching methods are generalizable to different environments~\cite{mehta_autonomous_2016}.

In short, existing works are mostly focused on modeling and replicating the detailed mechanisms of social compliance, which remains difficult to quantify due to the stochasticity in people's behaviors. In comparison, humans can intuitively evaluate whether a behavior is acceptable. In particular, human navigation (or teleoperation) is time-efficient and generally respects a set of simple social norms (e.g., ``passing on the right'')~\cite{kim_socially_2015,kretzschmar_socially_2016,knepper_pedestrian-inspired_2012}.
Building on a recent paper~\cite{chen_decentralized_2017}, we characterize these properties in a reinforcement learning framework, and show that human-like navigation conventions emerge from solving a cooperative collision avoidance problem.

The main contributions of this work are 
(i) introducing socially aware collision avoidance with deep reinforcement learning (SA-CADRL) for explaining/inducing socially aware behaviors in a RL framework, (ii) generalizing to multiagent ($n>2$) scenarios through developing a symmetrical neural network structure, and (iii) demonstrating on robotic hardware autonomous navigation at human walking speed in a pedestrian-rich environment.

\section{Background}\label{sec:prob_formulation}

\subsection{Collision Avoidance with Deep Reinforcement Learning}
A multiagent collision avoidance problem can be formulated as a sequential decision making problem in a reinforcement learning framework~\cite{chen_decentralized_2017}. Let $\mathbf{s}_t, \, \mathbf{u}_t$ denote an agent's state and action at time $t$, and let $\tilde{\mathbf{s}_t}$ denote the state of a nearby agent. To capture the uncertainty in nearby agents' intents, the state vector is partitioned into observable and unobservable parts, that is $\mathbf{s}_t = [\mathbf{s}_t^o, \, \mathbf{s}_t^h]$.  Let the observable states be the agent's position, velocity, and radius (size), $\mathbf{s}^o = [p_x, \, p_y, \, v_x, \, v_y, \, r] \in \mathbb{R}^{5}$; let the unobservable states be the agent's intended goal position, preferred speed, and orientation, $\mathbf{s}^h = [p_{gx}, \, p_{gy}, \, v_{pref}, \, \psi] \in \mathbb{R}^{4}$; and let the action be the agent's velocity, $\mathbf{u}_t = \mathbf{v}_t$. It will be explained in \cref{sec:results} that the observable states can be readily obtained from sensor measurements. The objective is to develop a policy, $\pi: \left( \mathbf{s}_t, \, \tilde{\mathbf{s}_t}^o \right) \mapsto \mathbf{u}_t$, that minimizes the expected time to goal $\mathbb{E}[t_g]$ while avoiding collision with nearby agents, 
\begin{align}
\argmin_{\pi\left(\mathbf{s}, \, \tilde{\mathbf{s}}^{o}\right)} \quad &\mathbb{E}  \left[t_g | \mathbf{s}_0, \, \tilde{\mathbf{s}}^o_0,  \, \pi \right] \label{eqn:cost} \\ 
s.t. \quad & ||\mathbf{p}_t - \tilde{\mathbf{p}}_t||_2 \geq r + \tilde{r} \qquad \forall t
		\label{eqn:con_collision} \\ 
	 \quad & \mathbf{p}_{t_g} = \mathbf{p}_g \label{eqn:con_reach_goal} \\
	 \quad & \mathbf{p}_t = \mathbf{p}_{t-1} + \Delta t \cdot \pi( \mathbf{s}_{t-1}, \, \tilde{\mathbf{s}}^o_{t-1}) \nonumber \\
	 \quad & \tilde{\mathbf{p}}_t = \tilde{\mathbf{p}}_{t-1} + \Delta t \cdot \pi( \tilde{\mathbf{s}}_{t-1}, \, \mathbf{s}^o_{t-1}), 
	 	\label{eqn:con_kinematics}
\end{align}
where \cref{eqn:con_collision} is the collision avoidance constraint, \cref{eqn:con_reach_goal} is the goal constraint, \cref{eqn:con_kinematics} is the agents' kinematics, 
and the expectation in \cref{eqn:cost} is with respect to the other agent's unobservable states (intents) and policy. 

This problem can be formulated in a reinforcement learning (RL) framework by considering an agent's joint configuration with its neighbor, $\mathbf{s}^{jn} = \left[ \mathbf{s}, \; \tilde{\mathbf{s}}^o \right]$.
In particular, a reward function, $R_{col}(\mathbf{s}^{jn}, \, \mathbf{u})$, can be specified to reward the agent for reaching its goal and penalize the agent for colliding with others.
The unknown state-transition model, $P( \mathbf{s}^{jn}_{t+1},\, \mathbf{s}^{jn}_t \,| \, \mathbf{u}_t)$, takes into account the uncertainty in the other agent's motion due to its hidden intents ($\tilde{\mathbf{s}}^h$). Solving the RL problem amounts to finding the optimal value function that encodes an estimate of the expected time to goal, 
\begin{align}
V^*(\mathbf{s}_0^{jn}) &= \mathbb{E} \left[ \sum_{t=0}^T \gamma^{t \cdot v_{pref}} \, R_{col}(\mathbf{s}^{jn}_t, \pi^*(\mathbf{s}^{jn}_t)) \; | \; \mathbf{s}_0^{jn} \right],
\label{eqn:optimal_value}
\end{align}
where $\gamma\in[0,1)$ is a discount factor. The optimal policy can be retrieved from the value function, that is 
\begin{align}
& \pi^*(\mathbf{s}^{jn}_{t+1}) = \argmax_{\mathbf{u}} R_{col}(\mathbf{s}_{t}, \mathbf{u}) + \nonumber \\ 
& \qquad \quad \gamma^{\Delta t \cdot v_{pref}}\int_{\mathbf{s}_{t+1}^{jn}}P(\mathbf{s}^{jn}_{t}, \mathbf{s}^{jn}_{t+1}|\mathbf{u}) V^*(\mathbf{s}_{t+1}^{jn})d\mathbf{s}_{t+1}^{jn}. \label{eqn:optimal_policy}
\end{align}

A major challenge in finding the optimal value function is that the joint state $\mathbf{s}^{jn}$ is a continuous, high-dimensional vector, making it impractical to discretize and enumerate the state space. Recent advances in reinforcement learning address this issue by using deep neural networks to represent value functions in high-dimensional spaces, and have demonstrated human-level performance on various complex tasks~\cite{mnih-dqn-2015,silver_mastering_2016,mnih_asynchronous_2016}. While several recent works have applied deep RL to motion planning~\cite{sadeghi_cad$^2$rl:_2016,zhang_learning_2016}, they are mainly focused on single agent navigation in unknown static environments, and with an emphasis on computing control inputs directly from raw sensor data (e.g., camera images). In contrast, this work extends the collision avoidance with deep reinforcement learning framework (CADRL)~\cite{chen_decentralized_2017} to characterize and induce socially aware behaviors in multiagent systems.

\begin{figure}[t]
	\centering
	\begin{subfigure}{0.15\textwidth}
		\centering
		\includegraphics [trim=100 170 600 200, clip, width=1.0 \textwidth, angle = 0, page=3]{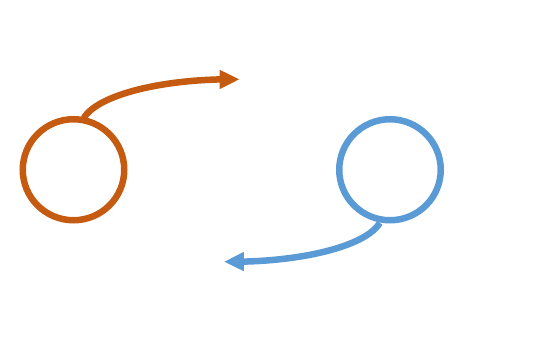}
	\end{subfigure}
	\begin{subfigure}{0.15\textwidth}
		\centering
		\includegraphics [trim=80 110 600 270, clip, width=1.0 \textwidth, angle = 0, page=5]{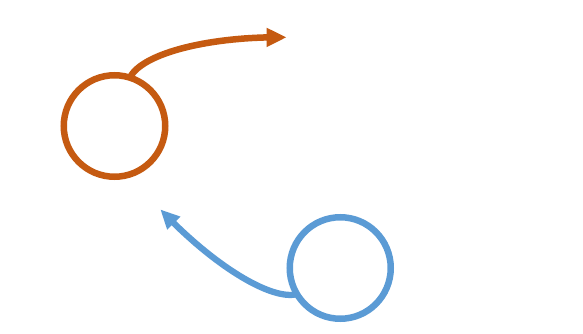}	
	\end{subfigure}
	\begin{subfigure}{0.15\textwidth}
		\centering
		\includegraphics [trim=100 170 600 220, clip, width=1.0 \textwidth, angle = 0, page=7]{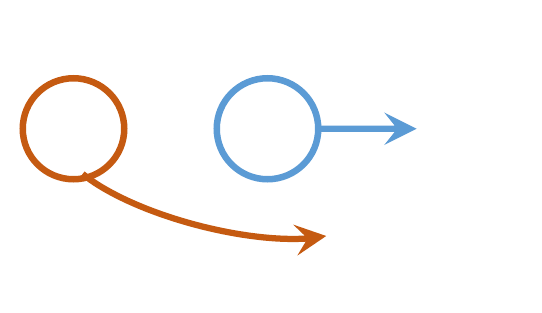}	
	\end{subfigure}
	\begin{subfigure}{0.15\textwidth}
		\centering
		\includegraphics [trim=100 196 600 230, clip, width=1.0 \textwidth, angle = 0, page=2]{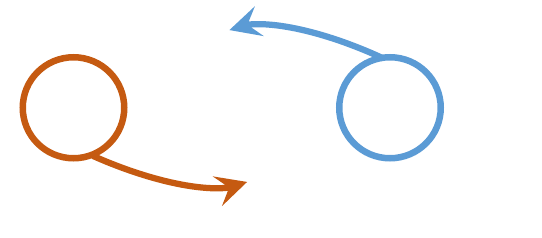}
		\caption*{passing}
	\end{subfigure}
	\begin{subfigure}{0.15\textwidth}
		\centering
		\includegraphics [trim=80 115 600 300, clip, width=1.0 \textwidth, angle = 0, page=4]{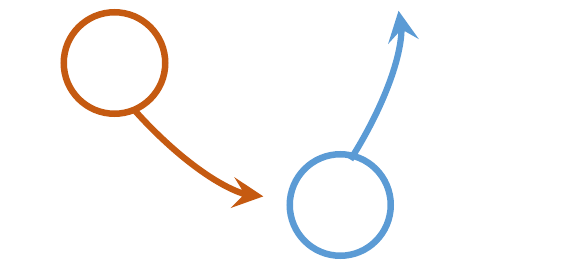}
		\caption*{crossing}
	\end{subfigure}
	\begin{subfigure}{0.15\textwidth}
		\centering
		\includegraphics [trim=100 201.5 600 220, clip, width=1.0 \textwidth, angle = 0, page=6]{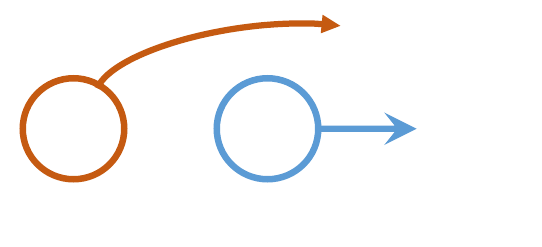}
		\caption*{overtaking}
	\end{subfigure}
	\caption{Symmetries in multiagent collision avoidance. Left to right show two equally time-efficient ways for the red agent to pass, cross and overtake the blue agent. The top row is often called the left-handed rules, and bottom row the right-handed rules.}
	\label{fig:symmetry}
\end{figure}

\subsection{Characterization of Social Norms} \label{sec:background:sn}
It has been widely observed that humans tend to follow simple navigation norms to avoid colliding with each other, such as passing on the right and overtaking on the left~\cite{knepper_pedestrian-inspired_2012}. Albeit intuitive, it remains difficult to quantify the precise mechanisms of social norms, such as when to turn and how much to turn when passing another pedestrian; and the problem exacerbates as the number of nearby pedestrians increases. This is largely due to the stochasticity in people's motion (e.g., speed, smoothness), which can vary significantly among different individuals~\cite{kretzschmar_socially_2016}.
\begin{figure}[t]
	\centering
	\begin{subfigure}{0.22\textwidth}
		\centering
		\includegraphics [trim=0 0 0 0, clip, width=1.0 \textwidth, angle = 0]{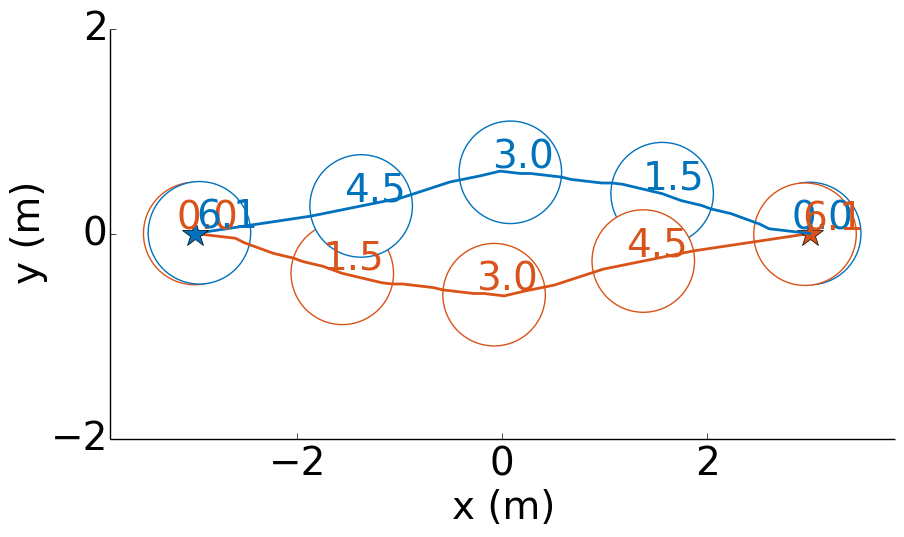}
		\caption{nominal case}
		\label{fig:cadrl_con_a} 
	\end{subfigure}
	\begin{subfigure}{0.22\textwidth}
		\centering
		\includegraphics [trim=0 0 0 0, clip, width=1.0 \textwidth, angle = 0]{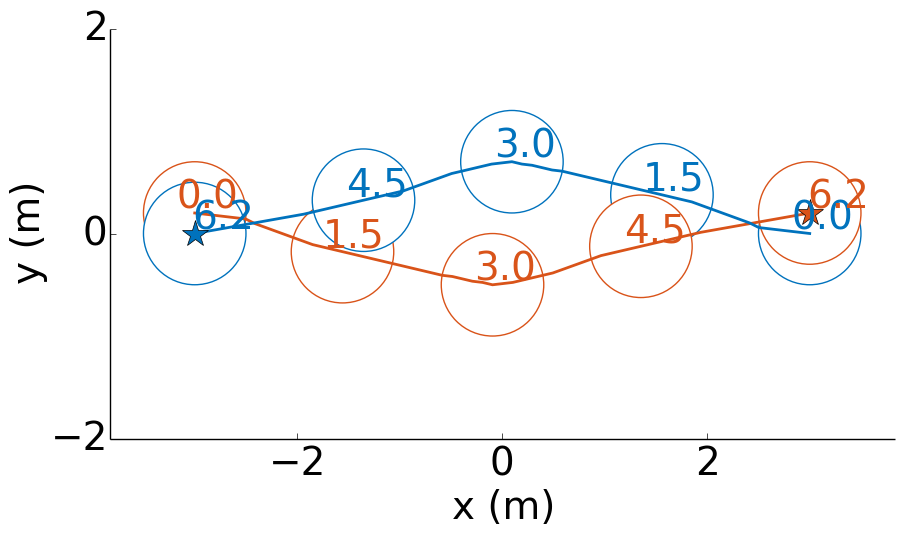}
		\caption{with 0.2m offset}
		\label{fig:cadrl_con_b} 
	\end{subfigure}
	\begin{subfigure}{0.22\textwidth}
		\centering
		\includegraphics [trim=0 0 0 0, clip, width=1.0 \textwidth, angle = 0]{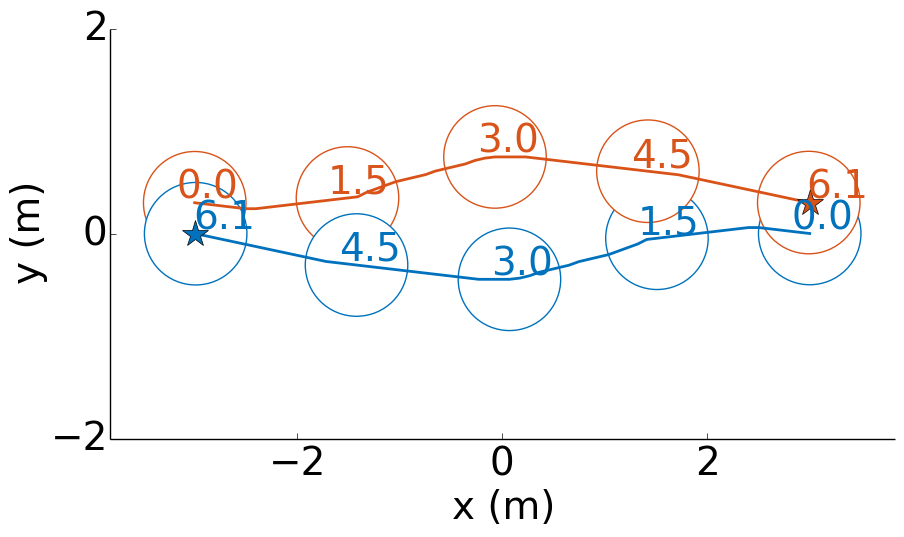}
		\caption{with 0.4m offset}
		\label{fig:cadrl_con_c} 
	\end{subfigure}
	\begin{subfigure}{0.22\textwidth}
		\centering
		\includegraphics [trim=0 0 0 0, clip, width=1.0 \textwidth, angle = 0]{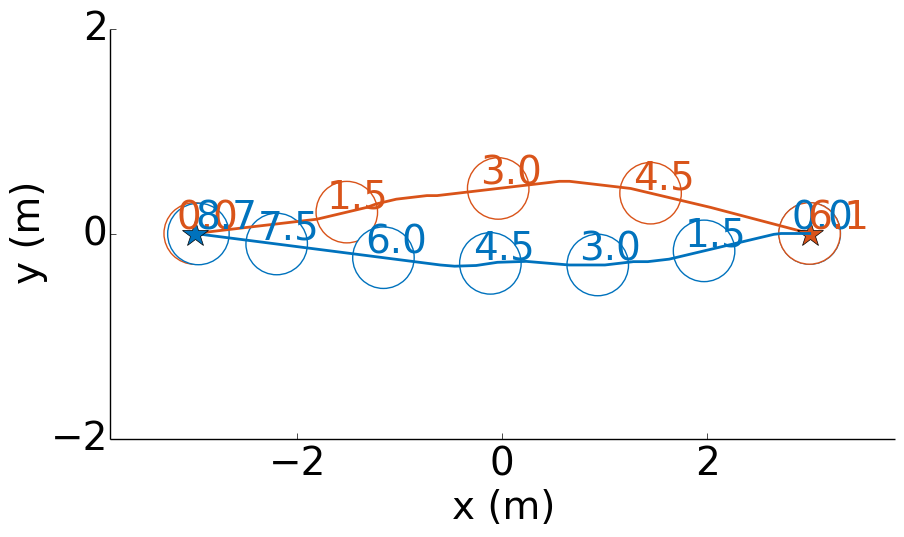}
		\caption{with diff. radii and speeds}
		\label{fig:cadrl_con_d} 
	\end{subfigure}
	\caption{Indications of a navigation convention from the CADRL policy. Circles show each agent's position at the labeled time, and stars mark the goals. (a) CADRL shows a preference to the right as two agents pass each other in a symmetrical test case (swapping position). (b) This passing direction is robust to a small offset in the initial condition. (c) The passing direction changes at a larger offset (balancing with time to reach the goal). (d) The convention is not consistent with human social norms, as two agents of different sizes and velocities exhibit a preference to the left.}
	\label{fig:cadrl_con}
\end{figure}

Rather than trying to quantify human behaviors directly, this work notes that the complex normative motion patterns can be a consequence of simple local interactions. For instance, an intuitive pairwise collision avoidance rule~\cite{helbing_social_1995} can cause simulated agents moving in the same direction to form lanes in long corridors. Thus, we conjecture that rather than a set of precisely defined procedural rules, social norms are the emergent behaviors from a time-efficient, reciprocal collision avoidance mechanism. Reciprocity implicitly encodes a model of the other agents' behavior, which is the key for enabling cooperation without explicit communication. Also, note that reciprocity does not require a unique set of navigation rules, since both the left-handed and the right-handed rules can resolve path conflicts as shown in~\cref{fig:symmetry}. Similarly, human navigation conventions are not unique, as the strength (e.g., separation distance) and passing direction vary in different countries~\cite{_wisdom_2011}.

Existing works have reported that human navigation (or teleoperation of a robot) tends to be cooperative and time-efficient~\cite{kim_socially_2015,kretzschmar_socially_2016}. This work notes that these two properties are encoded in the CADRL formulation through using the min-time reward function and the reciprocity assumption ($\tilde{\pi} = \pi$). Furthermore, it was interesting to observe that while no behavioral rules (e.g., function forms) were imposed in the problem formulation, CADRL policy exhibits certain navigation conventions, as illustrated in \cref{fig:cadrl_con}. In particular, \cref{fig:cadrl_con_a} illustrates two CADRL agents passing on the right of each other, showing signs of conforming to mutually agreed rules. More importantly, this preference in passing direction is robust to small deviations in the initial condition, as shown in \cref{fig:cadrl_con_b}. As the offset increases, the CADRL agents eventually change passing direction in favor of shorter, smoother paths (\cref{fig:cadrl_con_c}). Recall no communication took place and each agent's intent (e.g., goal) is not known to the other.

However, the cooperative behaviors emerging from a CADRL solution are not consistent with human interpretation. For instance, two CADRL agents with different sizes and preferred speeds show a preference to pass on the left of each other (\cref{fig:cadrl_con_d}). This is because an agent's state $\mathbf{s}$ is defined to be the concatenation of its position, velocity, size and goal, so it is unlikely that an emerging tie breaking navigation convention would solely depend on the relative position (as human social norms). Moreover, the cooperative behaviors of CADRL cannot be controlled -- they are largely dependent on the initialization of the value network and set of randomly generated training test cases. The next section will address this issue and present a method to induce behaviors that respect human social norms.

\section{Approach} \label{sec:approach}
The following presents the socially aware multiagent collision avoidance with deep reinforcement learning algorithm (SA-CADRL). We first describe a strategy for shaping normative behaviors for a two-agent system in the RL framework, and then generalize the method to multiagent scenarios.

\subsection{Inducing Social Norms}
\begin{figure}[t]
	\centering
	\includegraphics [trim=150 80 50 230, clip, width=0.5 \textwidth, angle = 0, page=3]{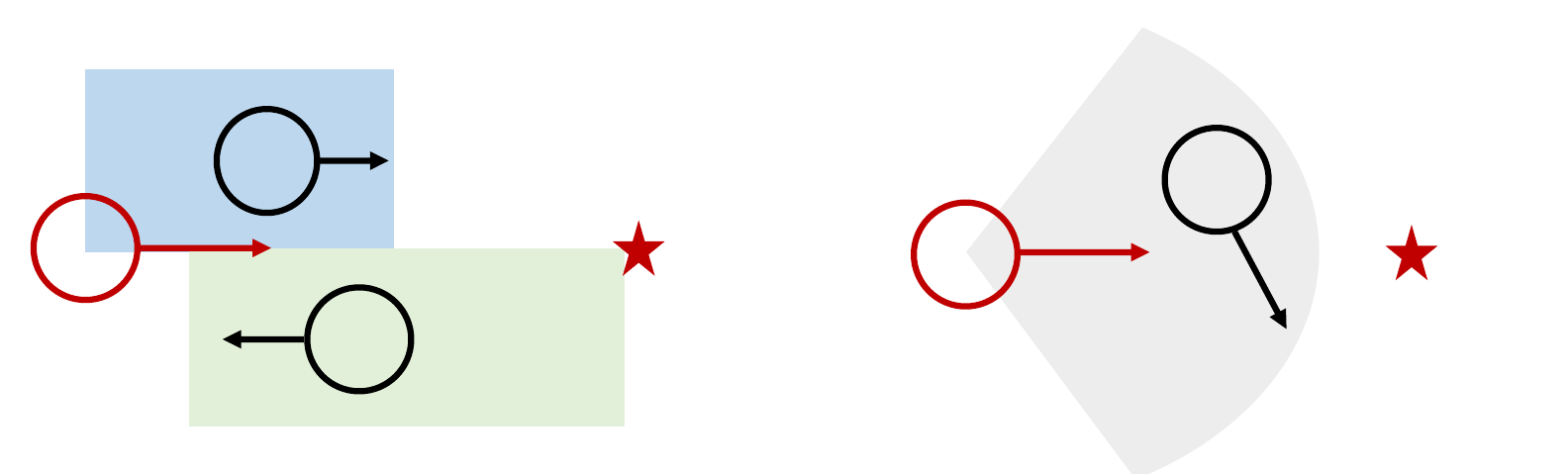}
	\caption{
	Norm inducing reward function (depiction of \cref{eqn:norm_pass}-\cref{eqn:norm_cross}). The red agent is penalized if there is another agent in the blue, green or gray shaded regions, corresponding to overtaking, passing and crossing, respectively. This induces the right-handed rules as shown in \cref{fig:symmetry}.}
	\label{fig:norm_reward}
\end{figure}

Recall the RL training process seeks to find the optimal value function \cref{eqn:optimal_value}, which maps from an agent's joint state with its neighbor, $\mathbf{s}^{jn} = \left[ \mathbf{s}, \; \tilde{\mathbf{s}}^o \right]$, to a scalar value that encodes the expected time to goal. To reduce redundancy (up to a rotation and translation), this work uses a local coordinate frame with the x-axis pointing towards an agent's goal, as shown in \cref{fig:norm_reward}. Specifically, each agent's state is parametrized as
\begin{align}
	\mathbf{s} & = [d_g, \; v_{pref}, \; v_x, \; v_y, \; \psi, \; r] \label{eqn:agent_state} \\  
	\tilde{\mathbf{s}}^o & = [\tilde{p}_x, \; \tilde{p}_y, \; \tilde{v}_x, \; \tilde{v}_y,
		 \; \tilde{r}, \; \tilde{d}_a, \; \tilde{\phi}, \; \tilde{b}_{on}] \;  \label{eqn:other_state}, 
\end{align}
where $d_g=||\mathbf{p}_g - \mathbf{p}||_2$ is the agent's distance to goal, $\tilde{d}_a=||\mathbf{p} - \tilde{\mathbf{p}}||_2$ is the distance to the other agent,  $\tilde{\phi}=\tan^{-1}(\tilde{v}_y / \tilde{v}_x)$ is the other agent's heading direction, and $\tilde{b}_{on}$ is a binary flag indicating whether the other agent is real or virtual (details will be provided in \cref{sec:approach:multi}).

This work notes that social norms are one of the many ways to resolve a symmetrical collision avoidance scenario, as illustrated in \cref{fig:symmetry}. To induce a particular norm, a small bias can be introduced in the RL training process in favor of one set of behaviors over others. For instance, to encourage passing on the right, states (configurations) with another agent approaching from the undesirable side can be penalized (green region in \Cref{fig:norm_reward}). The advantage of this approach is that violations of a particular social norm are usually easy to specify; and this specification need \emph{not} be precise. This is because the addition of a penalty breaks the symmetry in the collision avoidance problem, thereby favoring behaviors respecting the desired social norm. This work uses the following specification of a reward function $R_{norm}$ for inducing the right-handed rules (\cref{fig:norm_reward}),
\begin{align}
\hskip -0.25cm R_{norm}(\mathbf{s}^{jn}, \mathbf{u}) &= q_{n} I(\mathbf{s}^{jn} \in \mathcal{S}_{norm} \label{eqn:norm_cost}) \\ 
s.t. \quad \mathcal{S}_{norm} &= \mathcal{S}_{pass} \cup \mathcal{S}_{ovtk} \cup \mathcal{S}_{cross} \nonumber \\
 \quad \mathcal{S}_{pass} &= \{ \; \mathbf{s}^{jn} \; | \; d_g > 3, \quad 1<\tilde{p_x}<4, \nonumber \\                   & \quad -2<\tilde{p_y}<0, \quad |\tilde{\phi}-\psi| > 3\pi/4 \;  \} \label{eqn:norm_pass} \\
\quad \mathcal{S}_{ovtk} &= \{ \; \mathbf{s}^{jn} \; | \; d_g > 3, \quad 0<\tilde{p_x}<3, \quad |\mathbf{v}|>|\tilde{\mathbf{v}}| \nonumber \\                   &  \quad 0<\tilde{p_y}<1, \quad |\tilde{\phi}-\psi| < \pi/4 \;  \} \label{eqn:norm_overtake}  \\
\quad \mathcal{S}_{cross} &= \{ \; \mathbf{s}^{jn} \; | \; d_g > 3, \quad \tilde{d}_a < 2, \quad \tilde{\phi}_{rot} > 0, \nonumber \\                   &  \quad -3\pi/4<\tilde{\phi}-\psi < -\pi/4 \;  \} \; ,  \label{eqn:norm_cross}
\end{align}
where $q_{n}$ is a scalar penalty, $I(\cdot)$ is the indicator function, $\tilde{\phi}_{rot} = \tan^{-1}((\tilde{v}_x - v_x) / (\tilde{v}_y - v_y))$ is the relative rotation angle between the two agents, and the angle difference $\tilde{\phi}-\psi$ is wrapped between $[-\pi, \pi]$. An illustration of these three penalty sets is provided in \cref{fig:norm_reward}. 

\begin{figure}[t]
	\centering
	\begin{subfigure}{0.12\textwidth}
		\centering
		\includegraphics [trim=80 30 630 60, clip, width=1.0 \textwidth, angle = 0, page=2]{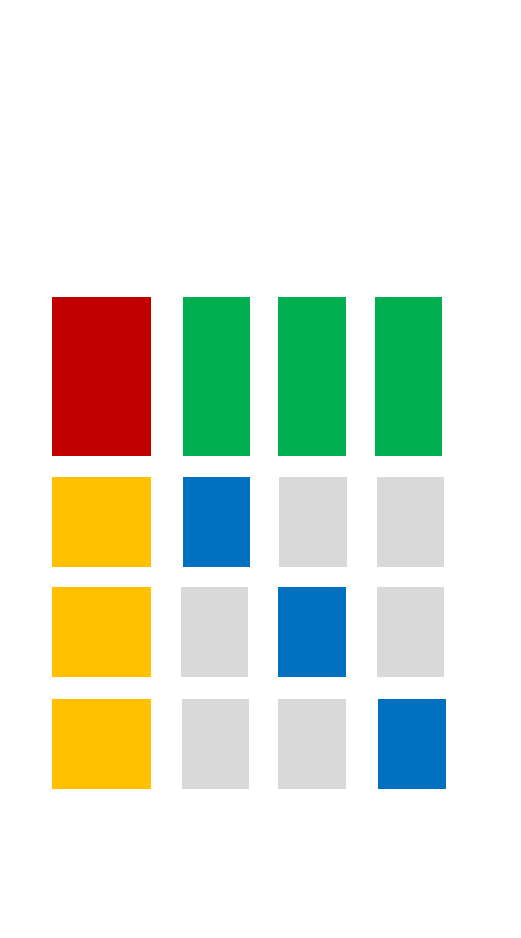}
		\caption{weight matrix}
			\label{fig:multinet_a}
	\end{subfigure}
	\hskip 0.02 \textwidth
	\begin{subfigure}{0.28\textwidth}
		\centering
		\includegraphics [trim=300 30 100 80, clip, width=1.0 \textwidth, angle = 0, page=2]{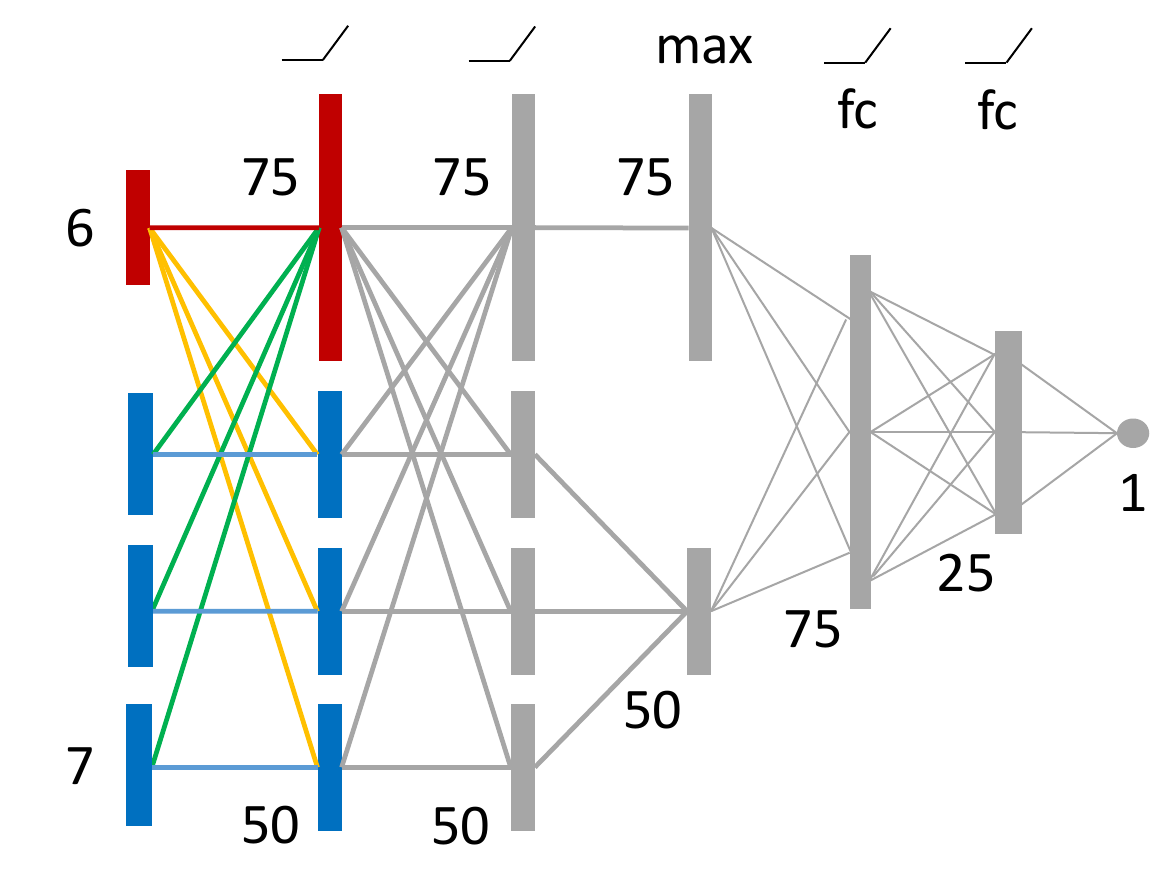}
		\caption{symmetric multiagent net}
			\label{fig:multinet_b}
	\end{subfigure}
	\caption{Network structure for multiagent scenarios. (a) illustrates the weight matrix used in the first layer, where block partitions with the same color are constrained to be the same. (b) shows the overall network structure, with two symmetric layers, one max-pooling layer, and two fully connected layers. The colored connections in the first layer correspond to (a), where red are the weights associated with the agent itself, and blue are that of the nearby agents.}
	\label{fig:multinet}
\end{figure}
\begin{figure*}[t]
	\centering
	\begin{subfigure}{0.99\textwidth}
		\centering
		\includegraphics [trim=0 0 0 0, clip, width=0.24 \textwidth, angle = 0]{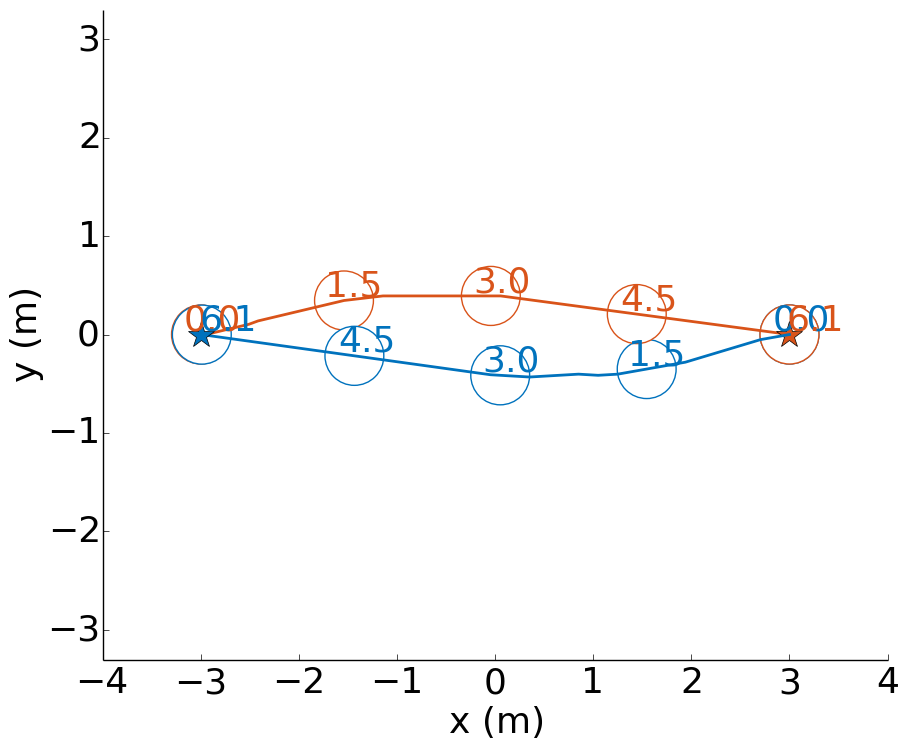}
		\includegraphics [trim=0 0 0 0, clip, width=0.24 \textwidth, angle = 0]{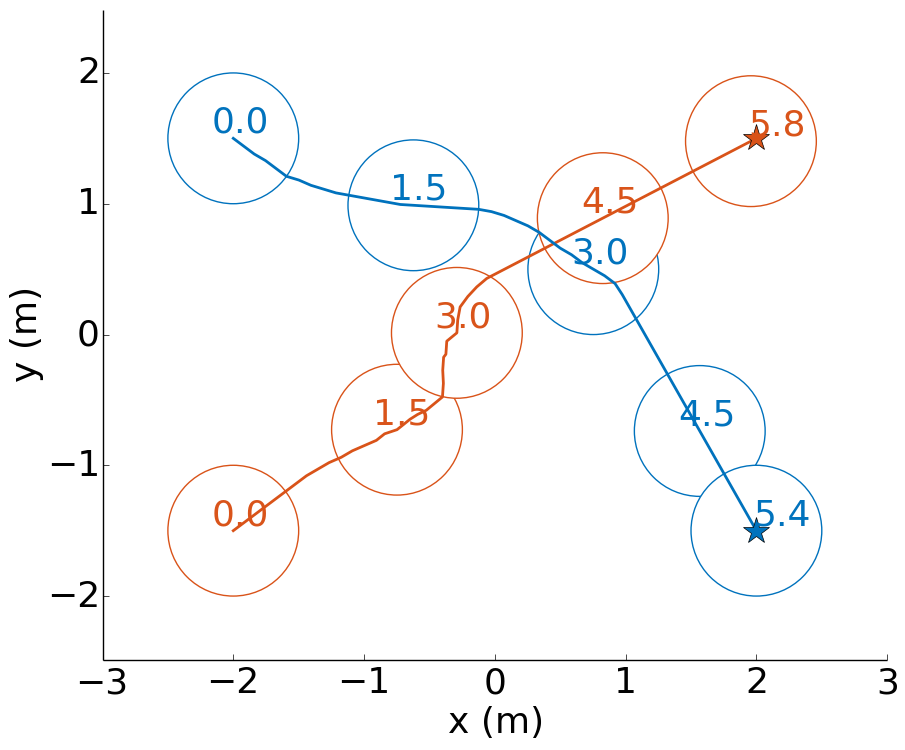}
		\includegraphics [trim=0 0 0 45, clip, width=0.24 \textwidth, angle = 0]{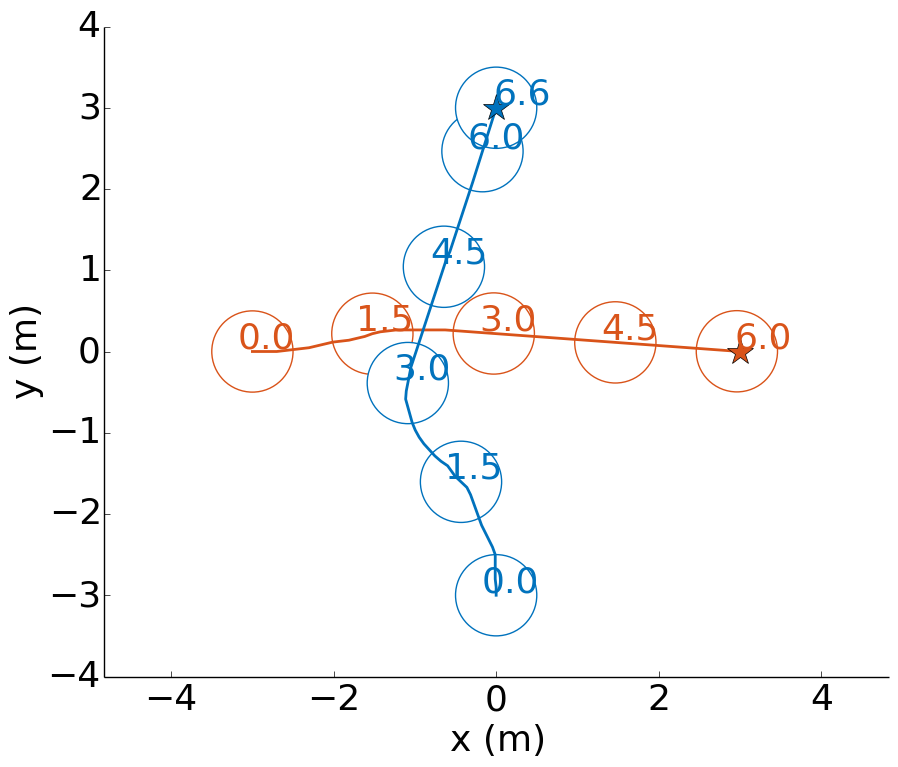}
		\includegraphics [trim=0 0 0 30, clip, width=0.24 \textwidth, angle = 0]{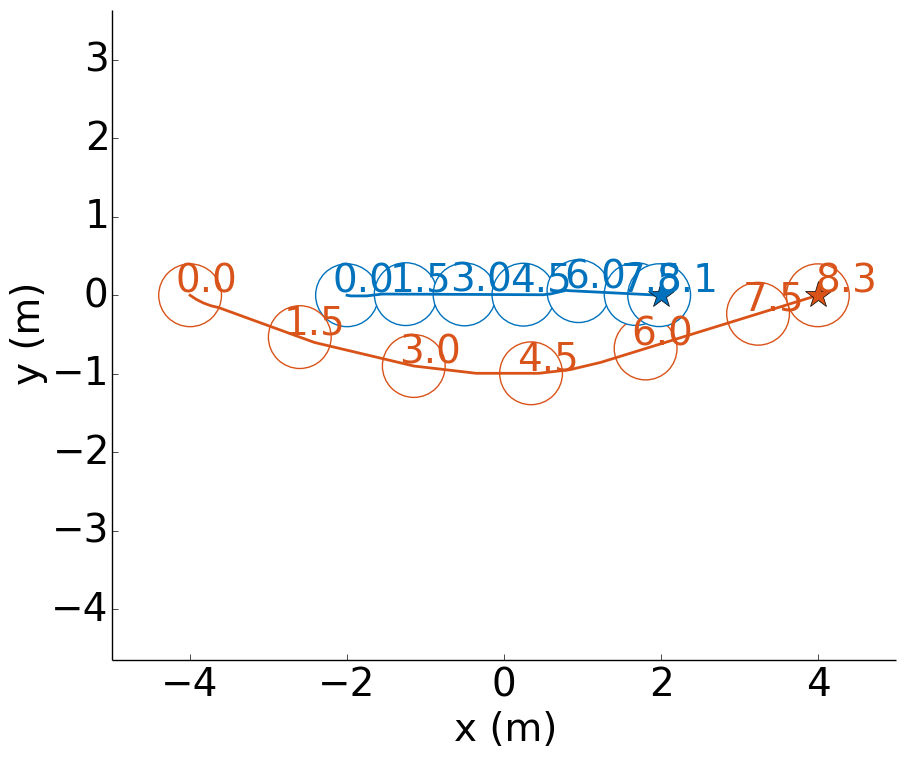}
		\caption{left-handed rule}
		\label{fig:sacadrl_traj_a} 
	\end{subfigure}
	\begin{subfigure}{0.99\textwidth}
		\centering
		\includegraphics [trim=0 0 0 0, clip, width=0.24 \textwidth, angle = 0]{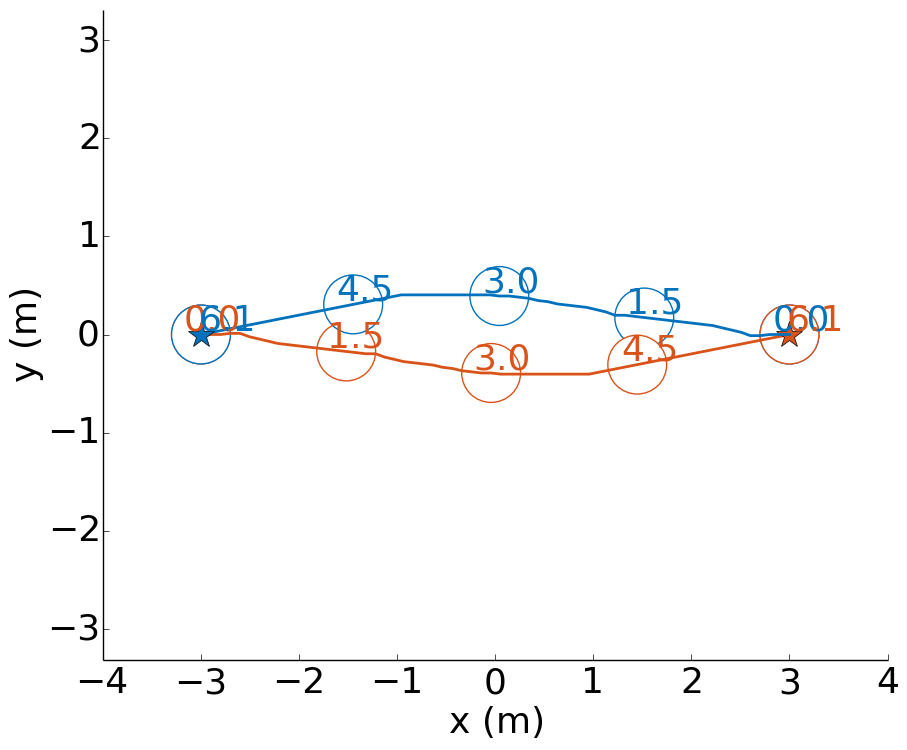}
		\includegraphics [trim=0 0 0 0, clip, width=0.24 \textwidth, angle = 0]{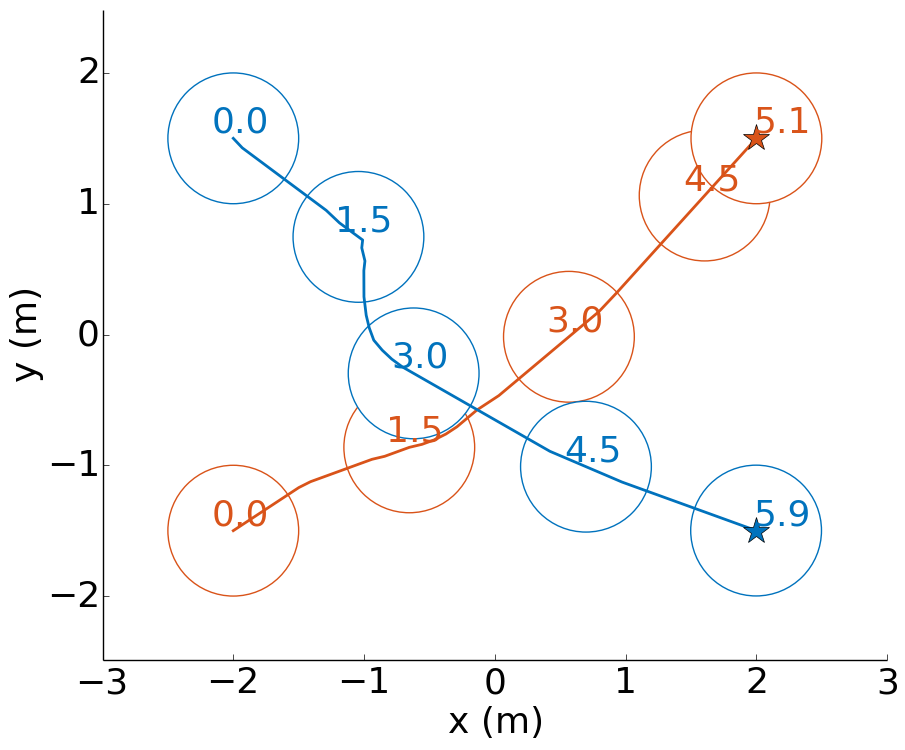}
		\includegraphics [trim=0 0 0 50, clip, width=0.24 \textwidth, angle = 0]{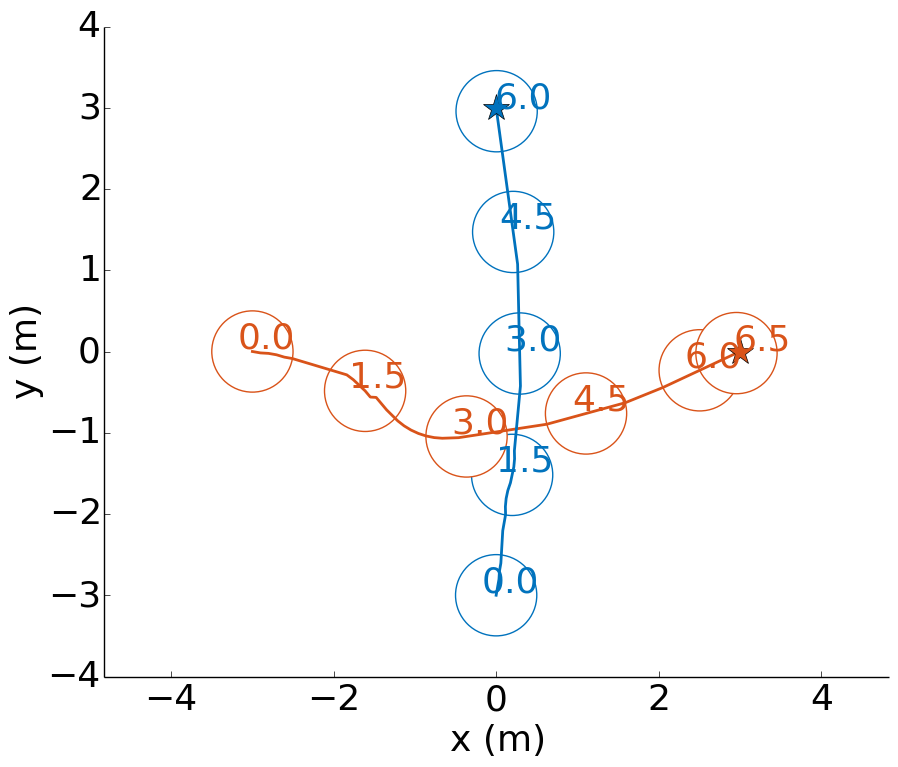}
		\includegraphics [trim=0 5 0 65, clip, width=0.24 \textwidth, angle = 0]{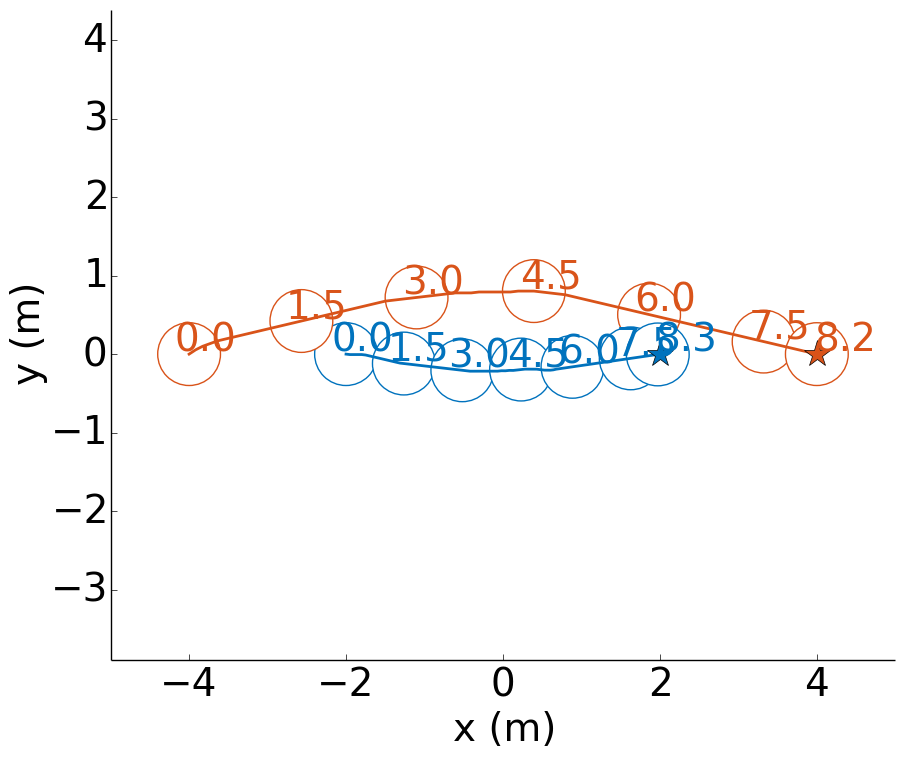}
		\caption{right-handed rule}
		\label{fig:sacadrl_traj_b}
	\end{subfigure}
	\caption{SA-CADRL policies exhibiting socially aware behaviors. Circles show each agent's position at the labeled time, and stars mark the goals. (a) and (b) show the trajectories generated by SA-CADRL policies trained to learn left-handed and right-handed rules, respectively. These behaviors are time-efficient and agree with the qualitative characterization of social norms shown in \cref{fig:symmetry}. Recall training is performed using randomly generated test cases.}
	\label{fig:sacadrl_traj}
\end{figure*}

The parameters defining the penalty set $\mathcal{S}_{norm}$ affect the rate of convergence. With \cref{eqn:norm_pass}-\cref{eqn:norm_cross}, the SA-CADRL policy converged within 700 episodes (exhibiting the desired behaviors such as passing on the right on all validation test cases). With a 30\% smaller penalty set (i.e., shrinking the shaded regions in \cref{fig:norm_reward}),
convergence occurred after 1250 episodes. Larger penalty sets, however, could lead to instability or divergence. Also, as long as training converges, the penalty sets' size does not have a major effect on the learned policy. This is expected because the desired behaviors are not in the penalty set. Similarly, \cref{eqn:norm_cost}-\cref{eqn:norm_cross} can be modified to induce left-handed rules. We trained two SA-CADRL policies to learn left-handed and right-handed norms starting from the same initialization, the results of which are shown in \cref{fig:sacadrl_traj}. The learned policies exhibited similar qualitative behaviors as shown in \cref{fig:symmetry}. Also note that training is performed on randomly generated test cases, and not on the validation test cases. 

\subsection{Training a Multiagent Value Network} \label{sec:approach:multi}
The CADRL work~\cite{chen_decentralized_2017} trained a two-agent network with three fully connected hidden layers, and used a minimax scheme for scaling up to multiagent ($n>2$) scenarios. Since training was solely performed on a two-agent system, it was difficult to encode/induce higher order behaviors, such as accounting for the relations between nearby agents. This work addresses this problem by developing a method that allows for training on multiagent scenarios directly.

To capture the multiagent system's symmetrical structure, a neural network with weight-sharing and max-pooling layers is employed, as shown in \cref{fig:multinet}. For a four-agent network shown in \cref{fig:multinet_b}, the three nearby agents' observed states can be swapped (blue input blocks) without affecting the output value.
This condition is enforced through weight-sharing, as shown in \cref{fig:multinet_a}. Two of such symmetrical layers are used,
followed by a max-pooling layer for aggregating features and two fully-connected layers for computing a scalar value. This work uses the rectified linear unit (ReLU) as the activation function in the hidden layers. 

The input to the n-agent network is a generalization of~\cref{eqn:agent_state}-\cref{eqn:other_state}, that is, $\mathbf{s}^{jn} = [\mathbf{s}, \; \tilde{\mathbf{s}}^{o,1}, \; \ldots \; \tilde{\mathbf{s}}^{o,n-1} ]$, where the superscripts enumerate the nearby agents. The norm-inducing reward function is defined similarly as \cref{eqn:norm_cost}, where a penalty is given if an agent's joint configuration with the \emph{closest} nearby agent belongs to the penalty set $\mathcal{S}_{norm}$. The overall reward function is the sum of the original CADRL reward and the norm-inducing reward, that is, $R(\cdot) = R_{col}(\cdot) + R_{norm}(\cdot)$. 

\begin{algorithm}[t]
	initialize and duplicate a value net with $n$ agents $V(\cdot;\mathbf{\theta},n)$, $ V' \leftarrow V$ \\
	initialize experience sets $E \leftarrow \emptyset$, $E_b \leftarrow \emptyset$ \\
	\For{episode={$1,\ldots, N_{eps}$}}{ 
		\For {m times}{
			$p$ $\sim$ Uniform(2, n) \\
			$\mathbf{s}_0^{1}, \mathbf{s}_0^{2}, \ldots, \mathbf{s}_0^{p}$ $\leftarrow$ randomTestcase(p) \\
			$\mathbf{s}_{0:t_f}^1,\mathbf{s}_{0:t_f}^{2}, \ldots, \mathbf{s}_{0:t_f}^{p} $ $\leftarrow$ SA-CADRL($V$) \\
			with prob $\epsilon_f$, mirror every traj $\mathbf{s}_{0:t_f}^i$ in the x-axis  \\
			\For {every agent i}{
				$y_{0:T}^i$ $\leftarrow$ findValues$\left( V', \; \mathbf{s}_{0:t_f}^{jn,i} \right)$ \\
				$E, \; E_b$ $\leftarrow$ assimilate$\left( E, \, E_b, \, (y^i, \mathbf{s}^{jn,i})_{0:t_f} \right)$ \\
			}
		}
		$e$ $\leftarrow$ randSubset($E$) $\cup$ randSubset($E_b$)\\
		$\mathbf{\theta}$ $\leftarrow$ RMSprop($e$) \\
		\For{every $C$ episodes}{
			 Evaluate($V$), $V'$ $\leftarrow$ $V$
		}
	}		
	\Return $ V $
	\caption{Deep V-learning for SA-CADRL}
	\label{alg:V-learning}
\end{algorithm}

The procedure for training a multiagent SA-CADRL policy is outlined in \cref{alg:V-learning}, which follows similarly as in~\cite{chen_decentralized_2017,mnih-dqn-2015}. A value network is first initialized by training on an n-agent trajectory dataset through neural network regression~(line~1). Using this value network \cref{eqn:optimal_policy} and following an $\epsilon$-greedy policy, a set of trajectories can be generated on random test cases (line 5-7). The trajectories are then turned into state-value pairs and assimilated into the experience sets $E, \, E_b$ (line 10-11). A subset of state-value pairs is sampled from the experience sets, and subsequently used to update the value network through back-propagation (line 12-13). The process repeats for a pre-specified number of episodes~(line~3-4).

Compared with CADRL~\cite{chen_decentralized_2017}, two important modifications are introduced in the training process. First, two experience sets, $E, \, E_b$, are used to distinguish between trajectories that reached the goals and those that ended in a collision (line~2,~11). This is because a vast majority ($\geq 90\%$) of the random generated test cases were fairly simple, requiring an agent to travel mostly straight towards its goal. The bad experience set $E_b$ improves the rate of learning by focusing on the scenarios that fared poorly for the current policy. Second, during the training process, trajectories generated by SA-CADRL are reflected in the x-axis with probability $\epsilon_f$ (line 8). By inspection of  \cref{fig:symmetry}, this operation flips the paths' topology (left-handedness vs right-handedness). Since a trajectory can be a few hundred steps long according to \cref{eqn:optimal_policy}, it could take a long time for an $\epsilon$-greedy policy to explore the state space and find an alternative topology. In particular, empirical results show that, without this procedure, policies can still exhibit the wrong passing side after 2000 training episodes. This procedure exploits symmetry in the problem to explore different topologies more efficiently. 

\begin{figure}[t]
	\centering
	\begin{subfigure}{0.48\textwidth}
		\centering
		\includegraphics [trim=0 0 0 0, clip, width=0.48 \textwidth, angle = 0]{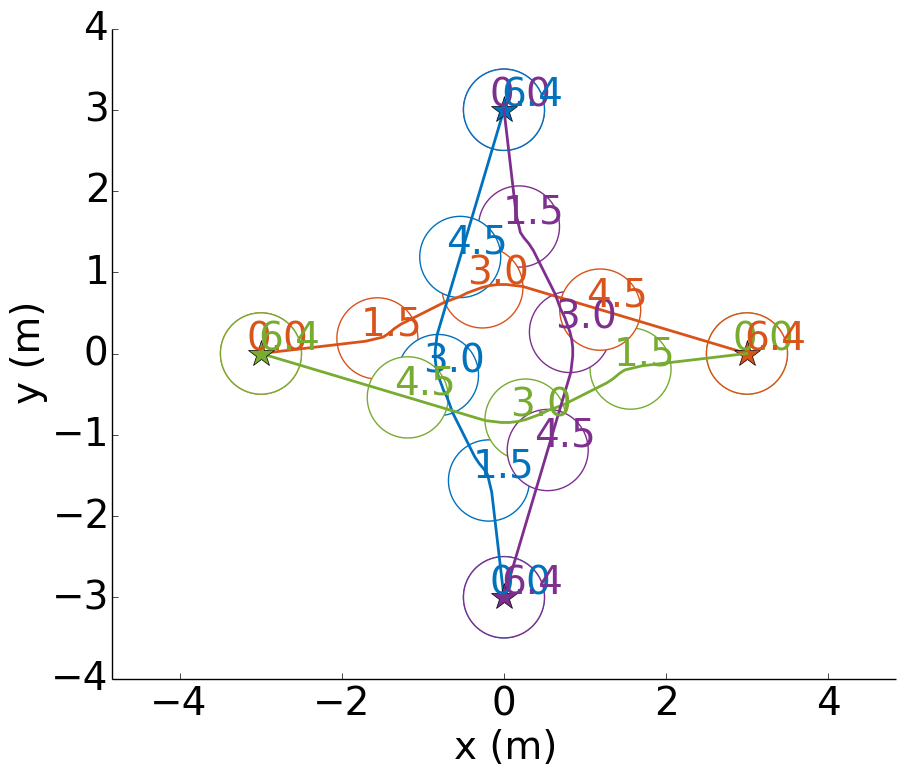}
		\includegraphics [trim=0 0 0 0, clip, width=0.48 \textwidth, angle = 0]{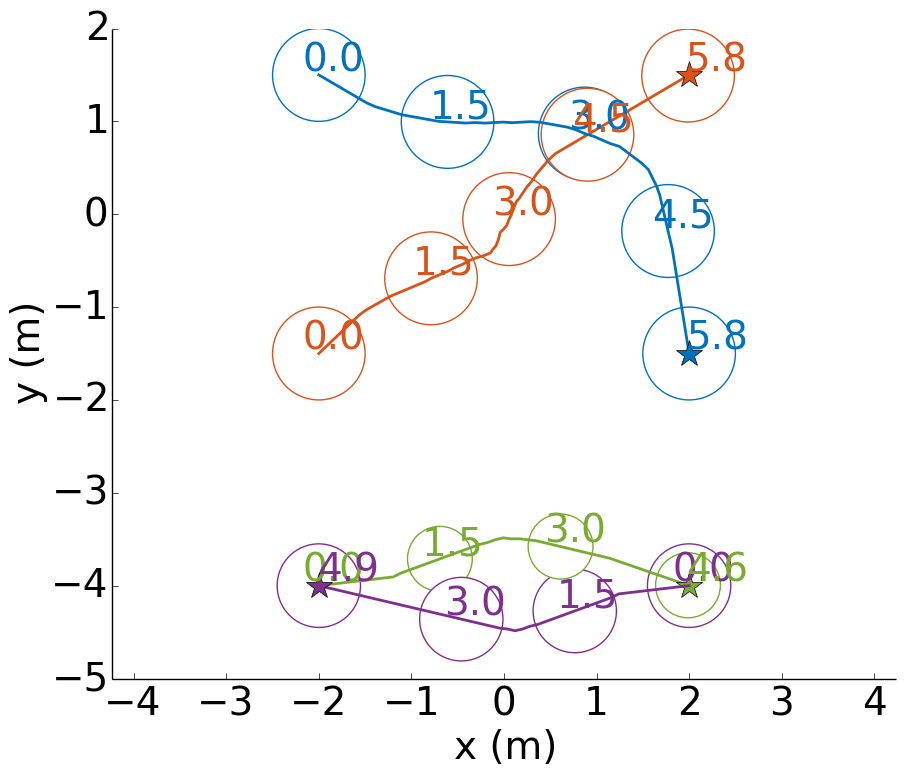}
		\caption{left-handed rule}
		\label{fig:sacadrl_multi_a} 
	\end{subfigure}
	\begin{subfigure}{0.48\textwidth}
		\centering
		\includegraphics [trim=0 0 0 0, clip, width=0.48 \textwidth, angle = 0]{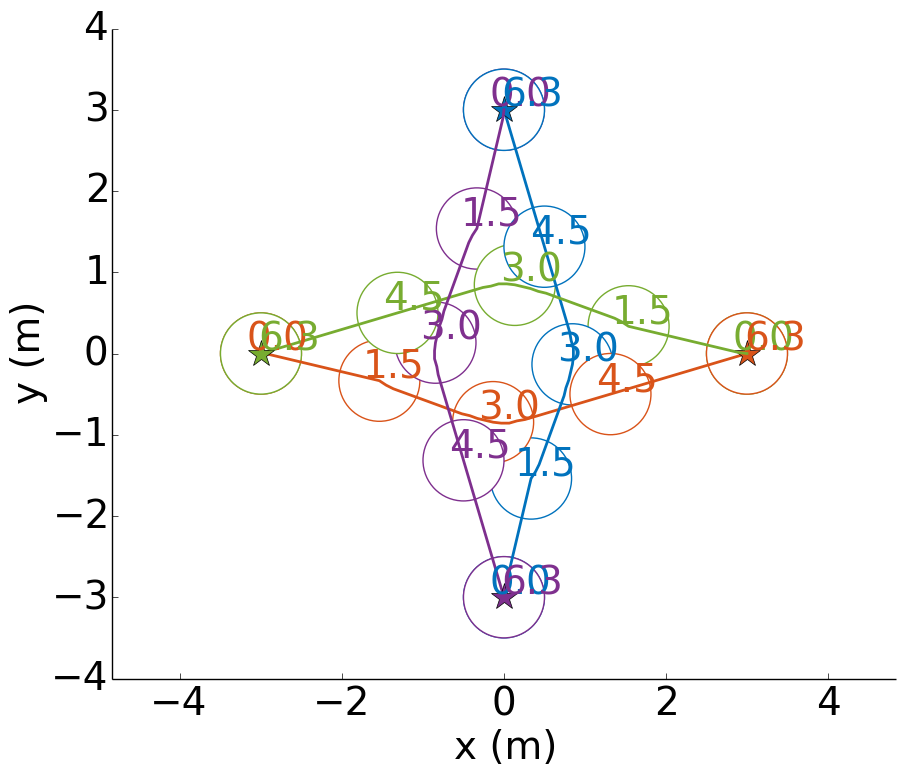}
		\includegraphics [trim=0 0 0 0, clip, width=0.48 \textwidth, angle = 0]{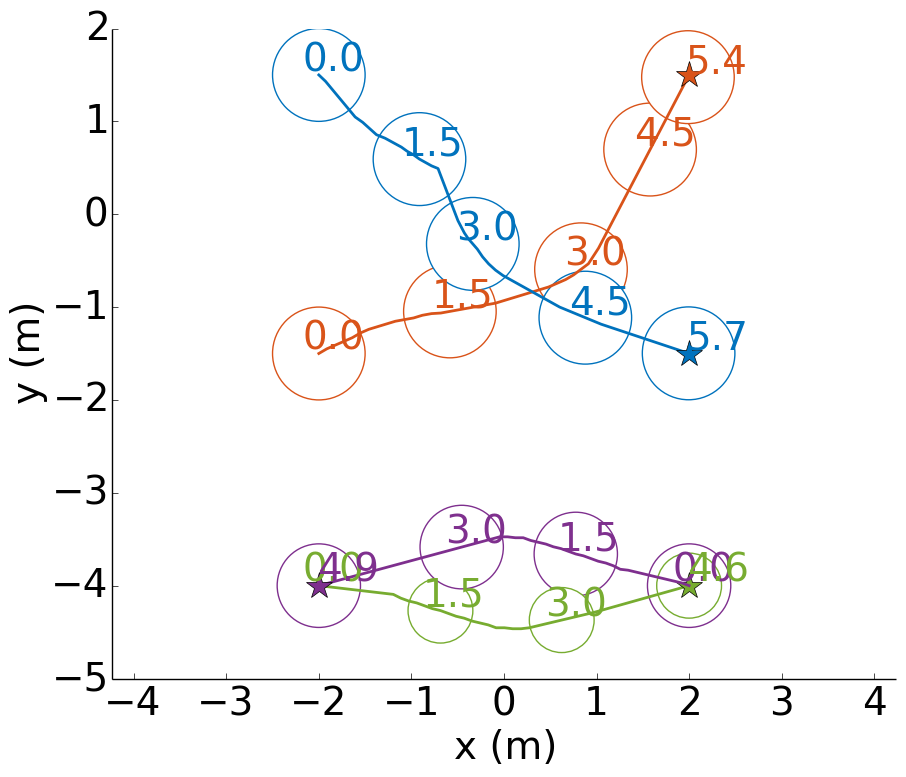}
		\caption{right-handed rule}
		\label{fig:sacadrl_multi_b}
	\end{subfigure}
	\caption{SA-CADRL policies generalized to multiagent scenarios using the network structure in \cref{fig:multinet}. Circles show each agent's position at the labeled time, and stars mark the goals. (a) and (b) show trajectories corresponding to the left-handed and right-handed rules, respectively.}
	\label{fig:sacadrl_multi} 
\end{figure}

Furthermore, an $n$-agent network can be used to generate trajectories for scenarios with fewer agents (line 5). In particular, when there are $p \leq n$ agents, the inputs in~\cref{fig:multinet_b} corresponding to the non-existent agents\footnote{Consider an agent with two nearby agents using a four-agent network.} can be filled by adding virtual agents -- replicating the states of the closest nearby agent and set the binary bit $\tilde{b}_{on}$ to zero \cref{eqn:other_state}. The use of this parametrization avoids the need for training many different networks. A left-handed and a right-handed four-agent SA-CADRL policies are trained using the network structure shown in \cref{fig:multinet}. Sample trajectories generated by these policies are shown in \cref{fig:sacadrl_multi}, which demonstrate the preferred behaviors of each respective set of social norms.

	\section{Results} \label{sec:results}
\subsection{Computational Details}
The size and connections in the multiagent network shown in \cref{fig:multinet} are tuned to obtain good performance (ensure convergence and produce time-efficient paths) while achieving real-time performance. In particular, on a computer with an i7-5820K CPU, a Python implementation of a four-agent SA-CADRL policy takes on average 8.7ms for each query of the value network (finding an action). Furthermore, offline training (\cref{alg:V-learning}) took approximately~nine hours to complete 3,000 episodes. In comparison, a two-agent network took approximately two hours for 1,000 training episodes. The four-agent system took much longer to train because its state space is much larger (higher dimensional) than that of the two-agent system. The training process was repeated multiple times and all runs converged to a similar policy -- exhibiting the respective desired social norms on all test cases in an evaluation set. Furthermore, this work generated random test cases with sizes and velocities similar to that of normal pedestrians~\cite{gates_recommended_2006}, such that $r~\in~[0.2,\, 0.5]$m, and $v_{pref} \in [0.3, \, 1.8]$m/s. Also, a desired minimum separation of 0.2m is specified through the collision reward $R_{col}$, which penalizes an agent for getting too close to its neighbors.

\subsection{Simulation Results}
Three copies of four-agent SA-CADRL policies were trained, one without the norm inducing reward $R_{norm}$, one with the left-handed $R_{norm}$, and the other with the right-handed $R_{norm}$. On perfectly symmetrical test cases, such as those shown in~\cref{fig:cadrl_con,fig:sacadrl_traj}, the left and right-handed SA-CADRL policies always select the correct passing side according to the respective norm. To demonstrate that SA-CADRL can balance between finding time-efficient paths and respecting social norms, these policies are further evaluated on randomly generated test cases. In particular, we compute the average extra time to reach the goals\footnote{$\bar{t}_e = \frac{1}{n} \sum_{i=1}^n [ t^i_{g} - ||\mathbf{p}^i_{0} - \mathbf{p}^i_{g}||_2 \; / \; v^i_{ pref}]$, where $t^i_{g}$ is the $i$th agent's time to reach its goal, and the second term is a lower bound of $t^i_{g}$ (straight toward goal at the preferred speed).}, the minimum separation distance, and the relative preference between left-handedness and right-handedness. Norm preference is calculated by counting the proportion of trajectories that violate the left-handed or the right-handed version of~\cref{eqn:norm_pass}-\cref{eqn:norm_cross} for more than 0.5 second. To ensure the test set is left-right balanced, each random test case is duplicated and reflected in the x-axis. Evidently, the optimal reciprocal collision avoidance (ORCA)~\cite{berg_reciprocal_2011} algorithm -- a reactive, rule-based method that computes a velocity vector based on an agent's joint geometry with its neighbors -- attains nearly 50-50 left/right-handedness on these test sets (first row of~\cref{tab:stat}).

The same four-agent SA-CADRL policies are used to generate trajectories for both the two-agent and the four-agent test sets.
On the two-agent test set, all RL-based methods produced more time-efficient paths than ORCA\footnote{ORCA specifies a reactive, geometric rule for computing a collision-free velocity vector, but it does not anticipate the evolution of an agent's state with respect to other agents nearby. Thus, ORCA can generate shortsighted actions and oscillatory paths (see~\cite{chen_decentralized_2017} for a detailed explanation). }. CADRL exhibited a slight preference to the right (40-60 split). The four-agent SA-CADRL (none) policy, in comparison, exhibited a stronger preference than ORCA and CADRL in each of the passing, crossing, and overtaking scenarios (third row in \cref{tab:stat}). This observation suggests that (i) certain navigation conventions could emerge as a means of resolving symmetrical collision avoidance scenarios, and (ii) the conventions don't always correspond to human social norms. For instance, SA-CADRL (none) prefers passing on the right but also overtaking on the right, which is a mix between right-handed and left-handed rules. In contrast, the SA-CADRL policies trained with $R_{norm}$ exhibited a strong preference (85-15 split) of the respective social norm. Recall that this ratio is not 1 because there is a tradeoff between time-optimality and social compliance, as illustrated in \cref{fig:cadrl_con}. This tradeoff can be controlled by tuning $q_n$ in~\cref{eqn:norm_cost}. Evidently, SA-CADRL (lh/rh) achieves better social compliance at a cost of an approximately 20\% larger $\bar{t}_e$, because satisfying the norms often requires traveling a longer path.

Similarly, the bottom rows of~\cref{tab:stat} show that in the four-agent test set, all RL-based methods outperformed ORCA, and SA-CADRL (lh/rh) exhibited behaviors that respect the social norms. CADRL produced paths that are closer to time-optimal than the other algorithms, but sometimes came very close (within 0.1m) to other agents. This close proximity occurred because CADRL was trained on a two-agent system, so its action choice is dominated by the single closest neighbor; possibly leading CADRL to select an action that avoids the closest neighbor but drives towards a third agent. In contrast, all SA-CADRL policies were trained on four-agent systems and they all maintained a larger average separation distance.

\begin{table*}[t]
	\centering
	\caption[]{SA-CADRL's performance statistics on 200 randomly generated test cases. SA-CADRL policies were trained (i) without the norm inducing reward $R_{norm}$, (ii) with left-handed $R_{norm}$, and (iii) right-handed $R_{norm}$, which are abbreviated as none, lh, rh, respectively. Results show that SA-CADRL policies produced time-efficient paths and exhibited behaviors that respect the corresponding social norm. } 
	\begin{tabular}{| c | c || c | c || K{1.5cm}| K{1.5cm} | K{1.5cm} ||}
	\hline
		Number of &
		\multicolumn{1}{c||}{\multirow{2}{*}{Method} \rule{0pt}{8pt} }  & \multicolumn{1}{c|}{\multirow{1}{*}{Extra time to goal $\bar{t}_e$(s) }} & 
		\multicolumn{1}{c||}{\multirow{1}{*}{Min separation dist. (m)}} &
		\multicolumn{3}{c||}{Norm preference (\%) [left-handed / right-handed]} 
		\\
		\cline{5-7}
		 \rule{0pt}{8pt} {agents} & {} & {[Avg / 75th / 90th pctl]} & { [10th pctl / avg]} & passing & crossing & overtaking \\ \hline
		\rule{0pt}{8pt} {\multirow{5}{*}{2}}  & {ORCA~\cite{berg_reciprocal_2011}} & 0.46 / 0.49 / 0.82 & 0.108 / 0.131 & 45 / 55 & 51 / 49 & 50 / 50 \\
		\rule{0pt}{8pt} {} & {CADRL~\cite{chen_decentralized_2017}}  & \textbf{0.25 / 0.30 / 0.47} & 0.153 / 0.189 & 37 / 63 & 38 / 62 & 43 / 57\\
		\rule{0pt}{8pt} {} & {SA-CADRL(none)}  & 0.27 / 0.28 / 0.54 & \textbf{0.169 / 0.189} & 10 / 90 & 32 / 68 & 63 / 37 \\
		\rule{0pt}{8pt} {} & {SA-CADRL(lh)}  & 0.30 / 0.36 / 0.67 & 0.163 / 0.192 & \textbf{98} / 2 &  \textbf{85} / 15 & \textbf{86} / 14\\
		\rule{0pt}{8pt} {} & {SA-CADRL(rh)} & 0.31 / 0.38 / 0.69 & 0.168 / 0.199 & 2 / \textbf{98} & 15 / \textbf{85} & 17 / \textbf{83} \\ \hline
		\rule{0pt}{8pt} {\multirow{5}{*}{4}} & {ORCA~\cite{berg_reciprocal_2011}} & 0.86 / 1.14 / 1.80 & 0.106 / 0.125 & 46 / 54 & 50 / 50 & 48 / 52 \\
		\rule{0pt}{8pt} {}  & {CADRL(minimax)~\cite{chen_decentralized_2017}}  & \textbf{0.41 / 0.54 / 0.76} & 0.096 / 0.173 & 31 / 69 & 41 / 59 & 46 / 54 \\
		\rule{0pt}{8pt} {} & {SA-CADRL(none)}  & 0.44 / 0.63 / 0.85 & \textbf{0.162 / 0.183} & 33 / 67 & 33 / 67 & 62 / 38\\
		\rule{0pt}{8pt} {} & {SA-CADRL(lh)}  & 0.49 / 0.69 / 1.00 & 0.155 / 0.178 &  \textbf{83} / 17 & \textbf{67} / 33 & \textbf{73} / 27 \\
		\rule{0pt}{8pt} {} & {SA-CADRL(rh)} & 0.46 / 1.63 / 1.02 & 0.155 / 0.180 & 12 / \textbf{88} & 29 / \textbf{71} & 30 / \textbf{70}\\ \hline
	\end{tabular} 
	\label{tab:stat}
	\vskip -0.1in
\end{table*}

\subsection{Hardware Experiment}
\begin{figure}[t]
	\begin{subfigure}{0.235\textwidth}
		\centering
		\includegraphics [trim=0 0 0 1, clip, width=1.0 \textwidth, angle = 0]{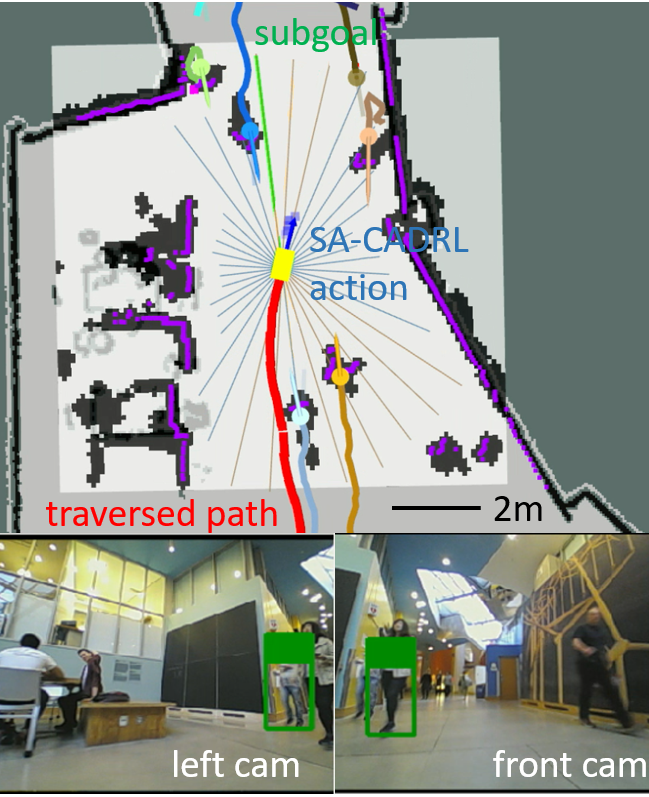}
		\caption{passing on the right}
		\label{fig:capture_a} 
	\end{subfigure}
	\hskip 0.01 \textwidth
	\begin{subfigure}{0.235\textwidth}
		\centering
		\includegraphics [trim=0 0 0 0, clip, width=1.0 \textwidth, angle = 0]{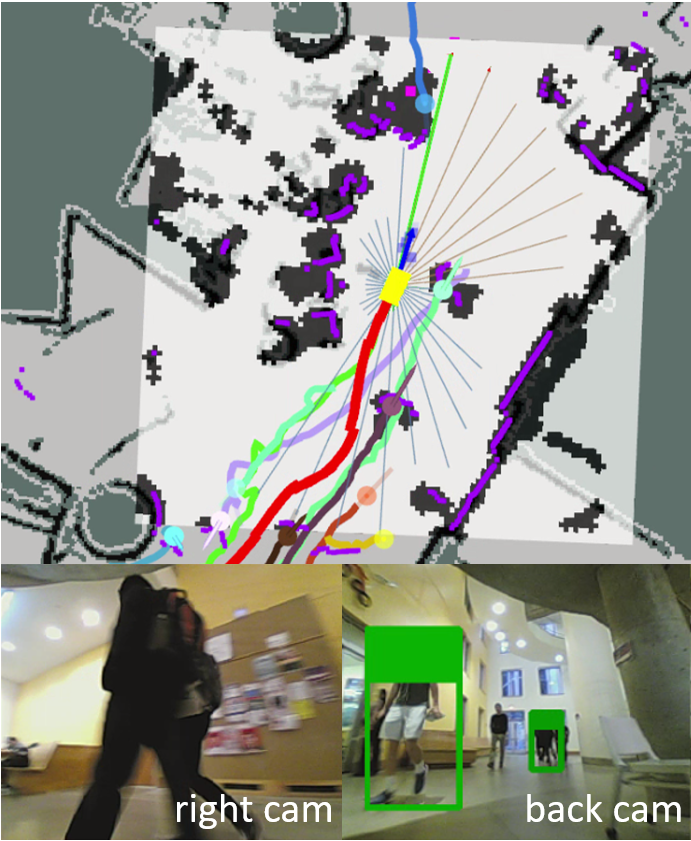}
		\caption{overtaking on the left}
		\label{fig:capture_b} 
	\end{subfigure}
	\caption[]{Visualization of a robotic vehicle's sensor data. The vehicle (yellow box) uses the right-handed SA-CADRL policy to navigate autonomously in an indoor environment at a nominal speed of 1.2m/s. (a) shows the vehicle passing a pedestrian on the right, where the person in the front camera image is detected and shown as the blue cylinder in the rviz view. (b) shows the vehicle overtaking a pedestrian on the left, where the person in the right camera image corresponds to the teal cylinder in the rviz view.}
	\label{fig:capture} 
	\vskip -0.1in
\end{figure}

The SA-CADRL policy is implemented on a robotic vehicle for autonomous navigation in an indoor environment with many pedestrians, as shown in~\cref{fig:rover}. The differential-drive vehicle is outfitted with a Lidar for localization, three Intel Realsenses for free space detection, and four webcams for pedestrian detection. Pedestrian detection and tracking is achieved by combining Lidar's pointcloud data with camera images~\cite{miller_dynamic_2016}. The speed, velocity, and size (radius) of a pedestrian are estimated by clustering the pointcloud data~\cite{campbell_dynamic_2013}. The estimated radius includes a buffer (comfort) zone as reported in~\cite{kim_socially_2015,kretzschmar_socially_2016}. Obstacles within a 10m $\times$ 10m square (perception range) centered at vehicle are detected and used to populate an occupancy map, shown as the white square in~\cref{fig:capture_a}. Interested readers are referred to~\cite{everett_robot_2017} for more details on the perception system and hardware construction of the robot. 

Motion planning uses the diffusion map algorithm~\cite{chen_motion_2016} for finding global paths and SA-CADRL for local collision avoidance. In particular, the diffusion map algorithm considers \emph{static} obstacles in the environment to find a subgoal within the vehicle's planning horizon (5m), which is shown at the end of the green line in~\cref{fig:capture_a}. Also, a set of feasible directions (heading and range) is computed, which corresponds to the free space in the occupancy map (white square). The feasible directions are visualized as the thin blue lines emanating from the vehicle. SA-CADRL takes in the set of detected pedestrians shown as the colored disks, and chooses an action (velocity vector) from the feasible directions to move the vehicle toward the subgoal. SA-CADRL's decision is shown as the blue arrow in \cref{fig:capture_a}, which does not necessarily line up with the subgoal. Note that pedestrians can be detected beyond the 5m static planning horizon, thus allowing socially aware interaction at a longer range. This whole sense-plan-execute cycle is fast enough to operate in real-time at 10Hz on a Gigabyte Brix computer onboard the vehicle.

Using this motion planning strategy, the vehicle was able to navigate fully autonomously in a dynamic indoor environment. In particular, the vehicle is issued randomly generated goals ten times, with an average distance between successive goals of more than 50 meters. During the experiment, an average of 10.2 persons came within 2m of the vehicle each minute. All encountered pedestrians are part of the regular daily traffic, and not testers/personnel associated with this work. For example, there were undergraduate students going between lectures and tourists visiting the campus. At a nominal speed of 1.2m/s, which is approximately the average human walking pace~\cite{gates_recommended_2006}, the vehicle maintained safe distance to the pedestrians and generally respected social norms. While a safety driver was walking approximately five meters behind vehicle, he never had to intervene or take over control at any time during the ten runs. Since people in North America follow the right-handed rule, the SA-CADRL policy with right-handed norms is used for the hardware experiment, which causes the vehicle to generally pass pedestrians on the right and overtake on the left. For examples, snippets of the experiment are shown in \cref{fig:capture}. A hardware demonstration video can be found at~\url{https://youtu.be/CK1szio7PyA}.

\section{Conclusion} \label{sec:conclusion}
This work presented SA-CADRL, a multiagent collision avoidance algorithm that considers and exhibits socially compliant behaviors. In particular, in a reinforcement learning framework, a pair of simulated agents navigate around each other to learn a policy that respect human navigation norms, such as passing on the right and overtaking on the left in a right-handed system. This approach is further generalized to multiagent ($n>2$) scenarios through the use of a symmetrical neural network structure. Moreover, SA-CADRL is implemented on robotic hardware, which enabled fully autonomous navigation at human walking speed in a dynamic environment with many pedestrians. Future work will consider the relationships between nearby pedestrians, such as a group of people who walk together.





\section*{Acknowledgment}
This work is supported by Ford Motor Company.
\balance
\bibliographystyle{IEEEtran} 
\bibliography{biblio}
\end{document}